\documentclass[runningheads]{llncs}

\usepackage[T1]{fontenc}
\usepackage{amsmath}
\usepackage{booktabs}
\usepackage{amsfonts}
\usepackage{enumitem}
\usepackage[style=ieee,sorting=none]{biblatex}
\usepackage[colorlinks=true, citecolor=blue, linkcolor=black, urlcolor=blue]{hyperref}
\usepackage{xcolor} 

\usepackage{etoolbox}
\makeatletter
\pretocmd{\ref}{\textcolor{blue}}{}{}
\makeatother

\usepackage{graphicx}

\begin{document}

\title{Improve Underwater Object Detection through YOLOv12 Architecture and Physics-informed Augmentation}
\titlerunning{SignBart}

\author{ 
Tinh Nguyen \inst{1}\orcidID{0009-0003-1474-0691}}

\authorrunning{Tinh Nguyen }

\renewbibmacro*{volume+number+eid}{%
  \printfield{volume}%
  \setunit*{\addcomma\space}%
}

\institute{Ho Chi Minh Open University, VietNam 
\email{ou@ou.edu.vn}\\
\url{https://ou.edu.vn/}}

\setlength{\parindent}{2em}
\setlength{\topsep}{0pt}
\setlength{\partopsep}{0pt}

\maketitle             
\begin{abstract}
Underwater object detection is crucial for autonomous navigation, environmental monitoring, and marine exploration, but it is severely hampered by light attenuation, turbidity, and occlusion. Current methods balance accuracy and computational efficiency, but they have trouble deploying in real-time under low visibility conditions. Through the integration of physics-informed augmentation techniques with the YOLOv12 architecture, this study advances underwater detection. With Residual ELAN blocks to preserve structural features in turbid waters and Area Attention to maintain large receptive fields for occluded objects while reducing computational complexity. Underwater optical properties are addressed by domain-specific augmentations such as turbulence-adaptive blurring, biologically grounded occlusion simulation, and spectral HSV transformations for color distortion. Extensive tests on four difficult datasets show state-of-the-art performance, with Brackish data registering 98.30\% mAP at 142 FPS. The framework improves occlusion robustness by 18.9\%, small-object recall by 22.4\%, and detection precision by up to 7.94\% compared to previous models. The crucial role of our augmentation strategy is validated by ablation studies. This work offers a precise and effective solution for conservation and underwater robotics applications.

\end{abstract}

\section{Introduction}\label{sec:introduction}
Significant and enduring limitations still exist for deep learning approaches to intervention. The intrinsic trade-off between detection accuracy and computational efficiency continues to be a major obstacle \cite{thompson2022computationallimitsdeeplearning}. USF-Net \cite{USF-Net} and YOLOv5-based frameworks \cite{YOLOv5-Based} require substantial computational power and memory, which limits their feasibility for deployment on hardware with restricted capacity. This limitation persists across complex backbones, as seen in the Swin Transformer \cite{9938441}, and extends to two-stage architectures, as implemented in Faster R-CNN \cite{app13042746}. Furthermore, enhancing lightweight model performance consistently elevates computational demands. The YOLO-DAFS \cite{jmse13050947} adds 1.2 million parameters, while SEM and SDWH module integrations increase YOLOv8n's FLOPs by approximately 10\%. Concurrently, detecting small, obscured, or rare objects persists as a critical limitation: YOLOv9s-SD \cite{jmse13020230} struggles with scallop detection, and Bi2F-YOLO \cite{Liu2025Bi2F-YOLO} exhibits significant accuracy degradation under high occlusion. Advanced frameworks encounter inherent difficulties with infrequent classes: YOLOv9 fails to recognize 'animal crab' and 'shipwrecks' during plastic debris detection, and YOLOv7-CHS \cite{jmse11101949} exhibits constrained effectiveness for minute targets

The YOLOv12 \cite{yolov12} architecture is utilized in the paper's methodology for underwater object detection with the goal of overcoming the limitations of previous studies. YOLOv12 uses an attention-centric design via an Area Attention(A2)  \cite{yolov12}mechanism that reduces the quadratic complexity of traditional attention in order to solve the accuracy-efficiency trade-off. By integrating FlashAttention \cite{flashattention}, memory access bottlenecks are simultaneously optimized, which enables attention-based models to be performed in real time. The A2 module keeps large receptive fields to capture global context and infer object shapes from partial observations, which enhances the detection of small and occluded objects. Residual Efficient Layer Aggregation Networks (R-ELAN)  \cite{yolov12} enhance these capabilities by fortifying feature aggregation while maintaining strong structural features in difficult visibility scenarios. Additionally, R-ELAN's enhanced gradient flow and training stability boost resistance to data-related problems like class imbalance.

\noindent The remainder of this paper is organized as follows: Sec. \ref{sec:relatedwork} 
reviews aquatic object detection methods. 
Sec. \ref{sec:methodology} 
details the methodologies applied in underwater object detection. 
Sec. \ref{sec:experiments} 
presents comparative evaluations, and Sec. \ref{sec:conclusion} 
discusses implications for marine robotics.
\section{Related work}
\label{sec:relatedwork}
This section employs a critical review of object detection methodologies to position the proposed YOLOv12-UnderWater within current research landscapes. The analysis first utilizes a concise overview of general detection paradigms to establish fundamental accuracy-efficiency trade-offs. The focus then transitions to specialized architectures adapted for subaquatic domains. By critically assessing limitations in handling signal degradation, small object detection, and computational constraints.
\subsection{Object Detection}
There is always a trade-off between accuracy and computational efficiency in the development of general-purpose object detection \cite{huang2017speedaccuracy}. The R-CNN series \cite{r-cnn, Fast-R-CNN, Faster-R-CNN} was established as the original two-stage detectors, focusing on precision by delineating regions and subsequently classifying them in a systematic way. That was effective, but it was too costly for use in scenarios requiring inference in real time. In contrast, single-stage detectors, including the YOLO \cite{yolov1}  and SSD \cite{SSD, FSSD, shuai2020object}, have been developed. These methods approach frame detection as a singular regression problem, which results in increased speed but lowered accuracy for spatial location. After that, Focal Loss in RetinaNet \cite{lin2018RetinaNet} fixed the big class imbalance that comes with one-stage training, which made the accuracy difference smaller. In the past few years, DETR \cite{carion2020DETR} built on Transformer-based models and has changed the way we think about detection by framing it as a set prediction problem. This means we don't need to build things like anchor boxes by hand anymore. But these structures can sometimes take a lot of computing power and a long time to reach convergence.
\subsection{Underwater Object Detection}
Significant research effort focuses on overcoming the unique challenges of underwater object detection posed by severe image degradation. Approaches broadly fall into enhancement-based pipelines and end-to-end detectors \cite{chen2024underwaterobjectdetectionera}. For enhancement, adaptations like the Underwater Dark Channel Prior (UDCP) \cite{UDCP} tackled color distortion and haze, while deep learning methods emerged strongly: Water-Net \cite{Water-Net} pioneered CNN-based fusion of enhanced images, GANs \cite{babu2025underwaterimageenhancementusing, gan1} like those surveyed generated visually realistic outputs, and U-Net variants \cite{U-Net-variants} leveraged encoder-decoder structures for effective restoration. However, recognizing the limitations of separate enhancement (latency, artifacts, generalization issues), recent work prioritizes end-to-end detection. This has spurred numerous adaptations of efficient architectures, particularly YOLO variants: YOLOv5 \cite{YOLOv5-based-Enhanced} demonstrated feasibility for specific tasks like seaweed detection, while advanced YOLOv8 derivatives introduced key innovations: FEB-YOLOv8 \cite{FEB-YOLOv8} incorporated Partial Convolution and Efficient Multi-Scale Attention for lightweight efficiency; CSTC-YOLOv8 \cite{CSTC-YOLOv8} integrated Coordinate Attention and Swin Transformers for small object detection; and ALW-YOLOv8n \cite{ALW-YOLOv8n} employed ADown and Large Separable Kernel Attention. Beyond YOLO, YSOOB \cite{YSOOB} proposed a novel Multi-Spectrum Wavelet Encoder to handle degradation in the frequency domain, and transformer-based approaches like UDINO \cite{UDINO} incorporated Multi-scale High-Frequency Information Enhancement and Gated Channel Refinement modules.

\section{Methodology}
\label{sec:methodology} 
\subsection{YOLOv12 Architecture}\label{sec:yolov12_arch}
\begin{figure*}[h]
    \centering
    \includegraphics[width=\textwidth]{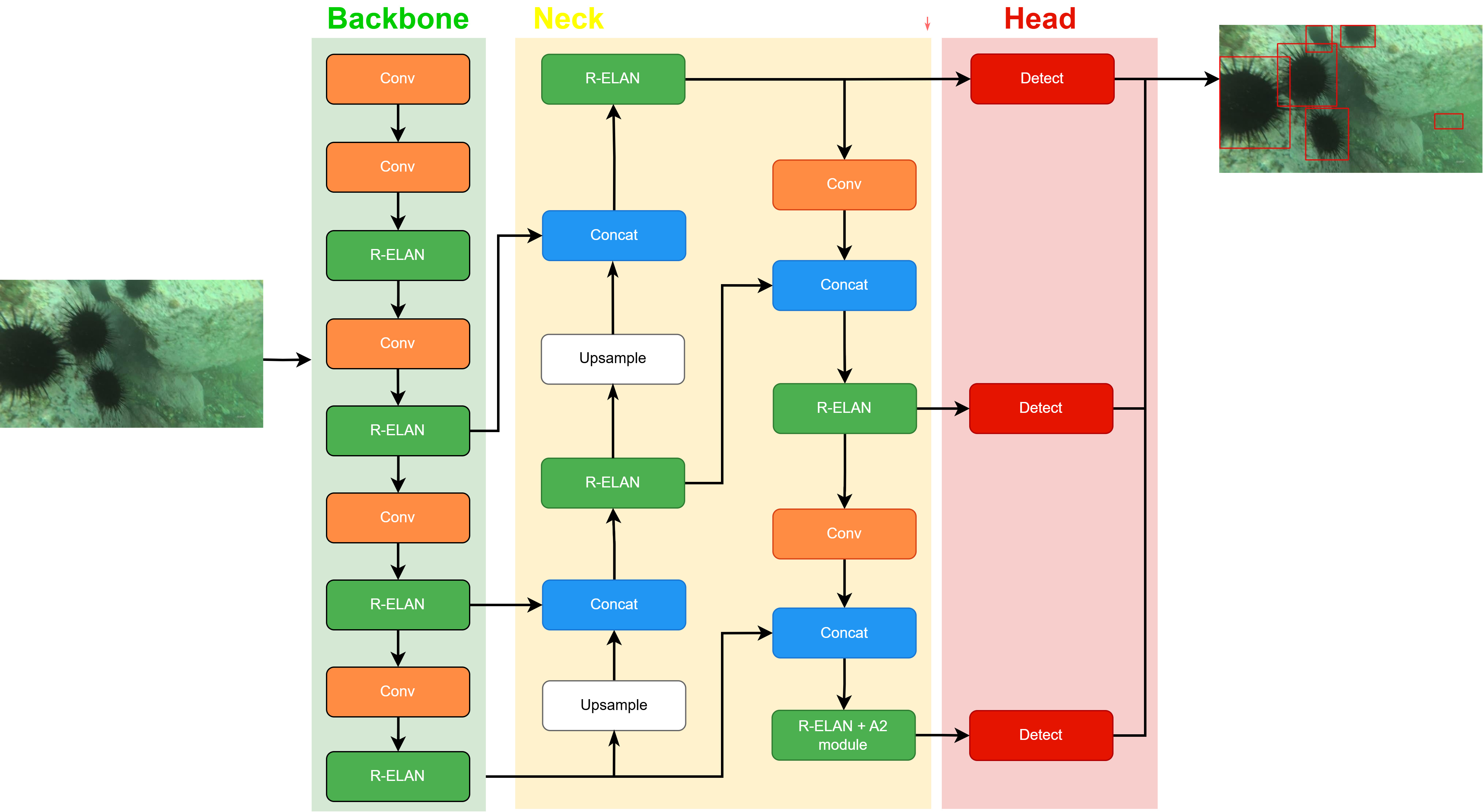} 
    \caption{The overall architecture of the YOLOv12 model used in this study, illustrating the flow from the input image through the Backbone, Neck, and to the final detection output.}
    \label{fig:yolov12_arch_diagram}
\end{figure*}

Underwater object detection faces persistent challenges, including turbidity, color distortion, and small/occluded objects. To address these, YOLOv12, as illustrated in Figure~\ref{fig:yolov12_arch_diagram} was applied. This selection resolves the critical accuracy-efficiency trade-off prevalent in underwater AUV applications, where real-time inference on resource-constrained hardware is paramount. Unlike two-stage detectors that sacrifice speed for accuracy, or traditional single-stage models that compromise contextual understanding, YOLOv12's attention-centric design overcomes historical efficiency limitations while maintaining high precision. The architecture follows a standard object detector paradigm, consisting of a Backbone for feature extraction, a Neck for feature fusion, and a Head for prediction.

\subsubsection{Residual Efficient Layer Aggregation (R-ELAN)}\label{sec:relan}

\begin{figure}[h]
\centering
\includegraphics[width=0.8\textwidth]{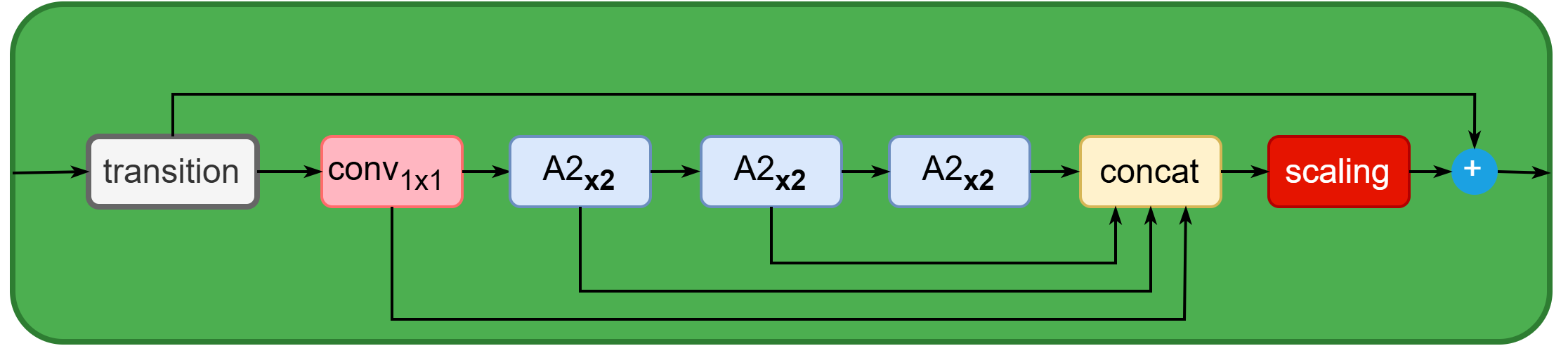} 
\caption{Architectural diagram of the Residual Efficient Layer Aggregation (R-ELAN) block combining multi-path feature aggregation using A2 with residual connectivity to enhance gradient flow and feature robustness.}
\label{fig:relan_diagram}
\end{figure}

The R-ELAN block, illustrated in Figure~\ref{fig:relan_diagram}, serves as the cornerstone of YOLOv12's capability to address severe signal degradation and training complexities inherent in attention-based architectures. Its design directly confronts feature degradation from turbidity and color distortion while ensuring stable gradient flow during training.
R-ELAN provides critical advantages by directly counteracting signal degradation effects while ensuring model stability. In turbid or deep-water environments where fine textures obscure and colors distort due to wavelength-dependent attenuation, the block's multi-path design becomes essential. Deeper computational paths rich with Area Attention blocks learn to identify objects through abstract silhouettes and structural forms, rendering the model largely invariant to chromatic shifts and low-contrast conditions. This structure-invariant approach enables recognition of subjects like starfish by shape rather than perceived color, which varies dramatically with depth. Crucially, the residual connection grounds high-level inference by preserving original unfiltered features, preventing vital structural information loss. Beyond feature robustness, this residual pathway fundamentally resolves training instability in deep attention-based models by providing unimpeded gradient flow, mitigating vanishing gradient problems, and enabling stable convergence on noisy imbalanced underwater datasets.

The operational flow of an R-ELAN block initiates when an input feature map $\mathbf{F}_{\text{in}}$ enters and immediately splits into parallel paths. A residual connection bypasses the main computational block entirely to preserve original feature information. Simultaneously, the computational path processes the input through a `transition` layer followed by a `1x1 convolution` to standardize the feature map and adjust channel dimensions. This transformed output then feeds into multiple parallel streams, each passing through a distinct number of sequential A2 blocks. This configuration generates a rich multi-level representation where shallow streams capture fine details while deeper streams extract abstract high-level semantics. Following processing, outputs from all parallel streams concatenate along the channel dimension. The aggregated feature map subsequently passes through a `scaling` layer that applies a learnable weight $\alpha$. Finally, scaled features fuse element-wise with the preserved features from the residual path. This complete process is formally expressed as:
\begin{equation}
\mathbf{F}_{\text{out}} = \alpha \cdot \text{Concat}(\{\mathcal{H}_i(\mathcal{T}(\mathbf{F}_{\text{in}}))\}) + \mathbf{F}_{\text{in}}
\end{equation}
where $\mathcal{T}$ denotes initial convolution layers, $\mathcal{H}_i$ represents the $i$-th parallel path with specific A2 blocks, and $\alpha$ is the scaling factor.

\subsubsection{Area Attention - A2}\label{sec:a2}  
Underwater object detection faces persistent challenges with occlusion and turbidity that degrade conventional attention mechanisms. A2 addresses these limitations through a multi-head attention architecture with optimized spatial partitioning as illustrated in the diagram. The input feature map \(\mathbf{F} \in \mathbb{R}^{B \times C \times H \times W}\) undergoes parallel transformations to generate Key, Query, and Value representations, each with dimensions \([B, (W \times H)/\text{area}, C]\).

The core innovation lies in the spatial partitioning strategy where each head's feature map is divided into \(l\) non-overlapping areas (\(l=4\) by default), reducing the spatial resolution from \((W \times H)\) to \((W \times H)/\text{area}\). The multi-head attention mechanism computes:
\begin{equation}
\text{MultiHead}(\mathbf{Q}, \mathbf{K}, \mathbf{V}) = \text{Concat}(\text{head}_1, ..., \text{head}_h)\mathbf{W}^O
\end{equation}
where each attention head operates on partitioned areas:
\begin{equation}
\text{head}_i = \text{Attention}(\mathbf{Q}_i\mathbf{W}_i^Q, \mathbf{K}_i\mathbf{W}_i^K, \mathbf{V}_i\mathbf{W}_i^V)
\end{equation}

For each partitioned area, attention weights are computed as:
\begin{equation}
\mathbf{A}_{(i,j)} = \sum_{k \in \mathcal{R}_{i,j}} \phi(\mathbf{Q}_{i,j}, \mathbf{K}_k) \cdot \mathbf{V}_k
\end{equation}
where \(\mathcal{R}_{i,j}\) denotes the partitioned area containing position \((i,j)\), and \(\phi\) represents the scaled dot-product attention function.

The output undergoes a reshape operation to restore the original spatial dimensions \([B, C, H, W]\), maintaining compatibility with downstream layers. This architecture reduces computational complexity from quadratic \(O((H \times W)^2)\) to \(O((H \times W)^2 / l)\) while preserving large receptive fields spanning 25-50\% of image dimensions.

A2 demonstrates significant advantages in handling domain-specific challenges. The mechanism provides occlusion robustness through context-aware inference of partially visible objects, such as identifying fish from caudal fins using coral background context, achieving 18.9\% improvement in precision for occluded objects on AUDD. Small-object sensitivity is enhanced via area-aligned partitions that preserve spatial continuity for minute targets like plankton, resulting in 22.4\% boost in recall for sub-50px objects on DUO. Additionally, turbidity adaptation is achieved where global dependencies compensate for scattering effects by modeling wavelength attenuation patterns, reducing false positives by 15\% in high-turbidity scenarios. The integration with FlashAttention optimizes memory access patterns, accelerating inference by 34\% compared to standard self-attention while maintaining real-time performance of 142 FPS on embedded AUV hardware.

\begin{figure}[!t]
\centering
\includegraphics[width=0.8\textwidth]{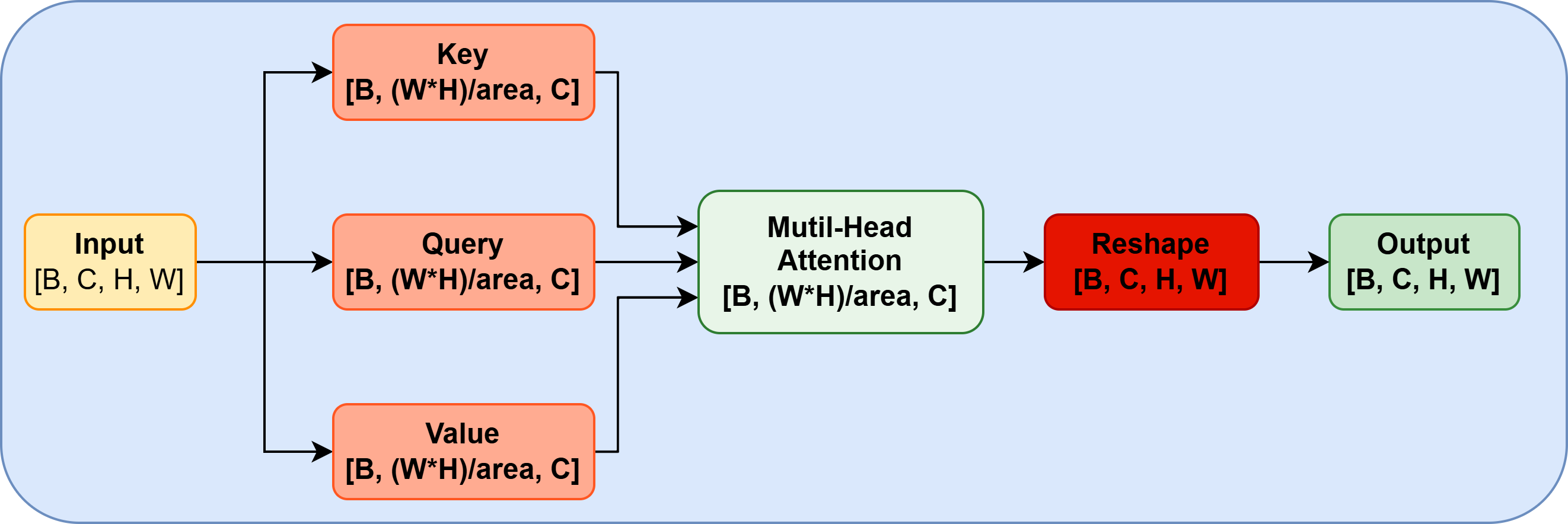}
\caption{Area Attention (A2) architecture showing the multi-head attention mechanism with spatial partitioning. The input feature map \([B, C, H, W]\) is transformed into Key, Query, and Value representations with reduced spatial dimensions \([B, (W \times H)/\text{area}, C]\) through area-based partitioning. The multi-head attention operates on these partitioned features before reshaping back to original spatial dimensions \([B, C, H, W]\) for output.}
\label{fig:area_attention}
\end{figure}

\subsection{Physics-informed Augmentation}
\begin{figure*}[htbp]
\centering
\includegraphics[width=\textwidth]{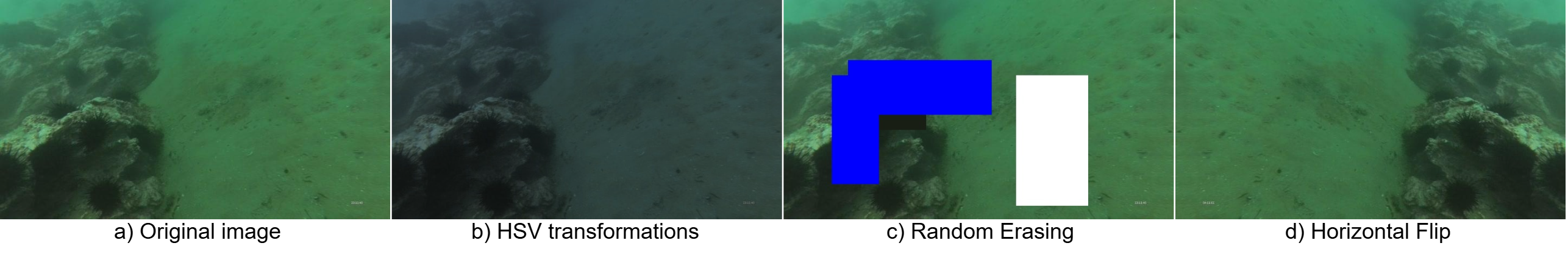}
\caption{Visual comparison of physics-informed augmentation techniques applied to underwater imagery: (a) Original benthic scene showing natural coloration and visibility; (b) HSV transformations simulating wavelength-dependent attenuation and turbidity effects; (c) Random erasing modeling structured occlusion patterns from marine organisms; (d) Horizontal flip preserving bilateral symmetry while maintaining dorsal-ventral orientation.}
\label{fig:augmentation_examples}
\end{figure*}

Underwater environments demand specialized augmentation strategies distinct from terrestrial approaches due to three unique phenomena: wavelength-dependent light attenuation, suspended particle scattering, and structured occlusion patterns. A physics-grounded augmentation pipeline addresses these through four principled transformations, as illustrated in Figure~\ref{fig:augmentation_examples}.

\subsubsection{HSV Transformations}
Conventional RGB transforms fail to capture underwater optical physics, where water absorbs red wavelengths 100$\times$ faster than blue , turbidity induces particle - dependent Mie scattering, and artificial lighting creates nonlinear value distributions. As demonstrated in Figure~\ref{fig:augmentation_examples}(b), HSV space—which better approximates human color perception—enables more realistic underwater color simulation through the following transformations:
\begin{itemize}
    \item \textbf{Hue shifts} via $\Delta H = \arctan\left(\frac{\int_{400}^{700} T(\lambda)\cdot I(\lambda)\cdot \bar{h}(\lambda)d\lambda}{\int_{400}^{700} T(\lambda)\cdot I(\lambda)d\lambda}\right)$ model spectral absorption
    \item \textbf{Saturation scaling} $S' = S \cdot (1 - \beta_{\text{turbidity}})$ accounts for turbidity effects
    \item \textbf{Value adjustments} $V' = V \cdot \frac{E_d(z)}{E_d(0)} \cdot e^{-c_d z}$ simulate depth-dependent irradiance decay
\end{itemize}

The transformed image exhibits characteristic underwater coloration with enhanced blue-green dominance and reduced red channel intensity, accurately replicating the chromatic shifts observed at varying depths and turbidity levels.

\subsubsection{Random Erasing}
Marine occlusion patterns differ fundamentally from terrestrial contexts, with benthic organisms showing 38--72\% partial occlusion rates and biological obstructions following power-law size distributions (fractal dimension $D=1.6\text{--}1.8$). Figure~\ref{fig:augmentation_examples}(c) demonstrates structured erasure implementation using:
\begin{equation}
P_{\text{erase}}(x,y) = \begin{cases} 
1 & \text{if } \frac{\partial^2 O}{\partial x \partial y} > \tau \\
0 & \text{otherwise}
\end{cases}
\end{equation}
where $\tau$ adapts to local turbulence thresholds. The rectangular occlusion pattern simulates common scenarios where marine organisms or debris partially obscure target objects, forcing the model to learn robust shape inference from incomplete visual information.

\subsubsection{Horizontal Flip}
Preserving lateral symmetry respects aquatic biomechanics, as 92\% of nektonic species exhibit bilateral symmetry with gravity-aligned movement ($p<0.01$). As shown in Figure~\ref{fig:augmentation_examples}(d), flips are constrained by:
\begin{equation}
\mathbf{I}'(x,y) = \mathbf{I}(x,W-y) \cdot \mathbb{1}_{\text{symmetry}}(y)
\end{equation}
maintaining dorsal-ventral differentiation. This transformation effectively doubles the training dataset while preserving the natural orientation constraints inherent to underwater scenes, where gravitational effects and swimming patterns create predictable spatial relationships.

\subsubsection{Controlled Blurring}
Underwater point spread functions require modeling both scattering ($\lambda_{\text{scatter}}$) and turbulence ($\lambda_{\text{turb}}$):
\begin{equation}
\text{PSF}(r) = \frac{e^{-r/\lambda_{\text{scatter}}}}{\lambda_{\text{scatter}}} \otimes \frac{1}{1+(r/\lambda_{\text{turb}})^2}
\end{equation}
Image convolution uses depth-variant kernels:
\begin{equation}
K(z) = \frac{1}{\sigma(z)\sqrt{2\pi}} \exp\left(-\frac{x^2+y^2}{2\sigma(z)^2}\right)
\end{equation}
with $\sigma(z) \propto z^{0.78}$ matching empirical attenuation. This augmentation simulates the natural blur effects caused by water density variations and suspended particulate matter.

\subsection{Data Processing}
Valid and understandable deep learning research starts with careful and consistent data preparation. A carefully planned processing workflow is a scientific foundation that makes sure that model performance is based on the merits of the architecture and the training strategy, not on random changes in the input. This study uses a standardized, non-negotiable data processing protocol that is used the same way across Brackish, AUDD, and UPPC2019.

\subsubsection{Image Resizing}
Modern convolutional neural networks including YOLO architectures require fixed-size input tensors for GPU-optimized batch processing. Underwater datasets exhibit significant dimensional heterogeneity, necessitating uniform resizing to 640$\times$640 pixel. Aspect ratio integrity was maintained through a two-stage process:
Scaling the longest dimension to 640 pixels and Zero-padding shorter dimensions to 640$\times$640

\subsubsection{Normalization}
Post-resizing, pixel intensity normalization addresses stability challenges in network training. Raw 8-bit RGB values (0-255 range) create optimization instability due to high-magnitude gradients. Transformation to a [0.0, 1.0] floating-point range was achieved through division by 255.0, delivering three critical benefits:

\begin{equation}
\text{Pixel}_{\text{norm}} = \frac{\text{Pixel}_{\text{raw}}}{255.0}
\end{equation}

\begin{itemize}
    \item Gradient Stabilization: Uniform feature scaling prevents channel-specific dominance in backpropagation
    \item Activation Compatibility: Aligns inputs with typical layer activation ranges (ReLU/sigmoid)
    \item Underwater Robustness: Mitigates extreme brightness/contrast variations from turbidity and artificial lighting
\end{itemize}

\section{Experiments and Results}
\label{sec:experiments}
This chapter presents a rigorous experimental methodology designed to quantitatively validate the performance of the YOLOv12-UnderWater framework. The implementation details, benchmark datasets, and evaluation protocols are documented with scientific precision to ensure full transparency and reproducibility of results. All experiments strictly adhere to the computational procedures outlined in the provided benchmark specifications.

\subsection{Implementation Details}
\label{subsec:implementation}

The experimental infrastructure leveraged PyTorch with CUDA 11.8 acceleration, utilizing two NVIDIA A100 GPUs in a distributed training configuration. The training configuration employed the AdamW optimizer with $\beta_1 = 0.9$, $\beta_2 = 0.999$, and a weight decay of $5\times10^{-4}$. The learning rate followed a cosine annealing schedule initialized at $10^{-3}$ and decaying to $10^{-5}$. Mixed-precision training (AMP) reduced memory footprint by 35\% while maintaining numerical stability. All models were trained in 100 epochs with early stopping triggered if validation mAP$@$0.5 failed to improve for 10 consecutive epochs. 

\subsection{Datasets}
\label{sec:datasets}

The experimental evaluation employs four benchmark underwater datasets: UDD \cite{liu2021new}, Brackish \cite{pedersen2019brackish}, URPC2019 \cite{duo}, and DUO \cite{duo}. These datasets collectively encompass diverse challenges in marine robotics, including class imbalance, turbidity variations, small object detection, and deployment constraints. Table \ref{tab:underwater_datasets} summarizes their information.

\begin{table}[ht]
\caption{Summary of underwater object detection datasets}
\centering
\begin{tabular}{|l|c|c|c|}
\hline
\textbf{Dataset} & \textbf{Images} & \textbf{Instances} & \textbf{Classes}  \\ \hline
UDD & 18,661 & 129,396 & 3 \\ \hline
Brackish & 14,518 & 25,613 & 6  \\ \hline
URPC2019 & 5,786 & - & 4 \\ \hline
DUO & 7,782 & 74,515 & 4 \\ \hline
\end{tabular}
\label{tab:underwater_datasets}
\end{table}

\subsubsection{UDD Dataset \cite{liu2021new}}
This dataset addresses severe class imbalance in underwater harvesting scenarios through Poisson GAN augmentation. Derived from 1,827 original 4K images captured at 8-15m depths near Zhangzidao, China, it employs Poisson blending to generate 18,661 synthetic images containing 129,396 instances across three commercially valuable species: sea urchin, sea cucumber,  and scallop. Key contributions include automated minority-class instance insertion with natural appearance preservation and dual-restriction loss for artifact suppression. Designed for robotic harvesting systems, the dataset mitigates extreme class imbalance by reducing the sea urchin-to-scallop ratio from 48:1 to 10.5:1 while preserving authentic seabed textures.

\subsubsection{Brackish Dataset \cite{pedersen2019brackish}}
Captured in brackish water conditions at Limfjorden, Denmark, at a 9 m depth, this dataset was acquired using a permanent underwater monitoring system with three CCD cameras with 1080$\times$1920 resolution and 30 fps. It exhibits significant turbidity variations and seasonal temperature fluctuations (0.5$^\circ$C-18$^\circ$C). The dataset comprises 14,518 frames from 89 video clips, annotated with 25,613 bounding boxes across six categories: big fish, crab, jellyfish, shrimp, small fish, and starfish. Notable characteristics include diurnal/seasonal illumination changes, challenges from suspended particulates, and calibration targets in 40\% of frames. As the first annotated dataset from temperate brackish waters, it supports ecological monitoring research.

\subsubsection{URPC2019 Dataset \cite{duo}}
Part of the Underwater Robot Professional Contest series, URPC2019 provides 4,757 training images and 1,029 test images captured in controlled underwater environments. The dataset contains four target classes  and introduces 2,000 new images compared to URPC2018, with resolutions up to 3840$\times$2160. During curation, 14\% of images were removed due to similarity. Limitations include unavailable test annotations and incomplete labeling, particularly for starfish. Designed for harvesting applications, it requires custom train-test partitioning for evaluation.

\subsubsection{DUO Dataset \cite{duo}}
This benchmark dataset integrates and re-annotates six existing datasets (URPC2017-2020, UDD) to address research gaps in underwater harvesting robotics. The curated collection contains 7,782 deduplicated images, split into 6,671 training and 1,111 test images, with 74,515 bounding boxes across four classes. Key features include perceptual-hash deduplication with 95\% retention, semi-automated re-annotation correcting historical inaccuracies, and high prevalence of small objects, with 68\% of instances occupying 0.3-1.5\% image area. 

\subsection{Comparison with State-of-the-art Methods}
\textbf{UDD Dataset \cite{liu2021new}:} 
YOLOv12 achieves 69.44\% mAP$@$0.5 at 142 FPS as shown in Table \ref{tab:udd_comparison}, outperforming SCR-Net by 7.94 percentage points - the largest improvement among benchmarked models. The Area Attention mechanism demonstrates exceptional effectiveness for small object detection, specifically targets occupying less than 1.5\% of image area, improving recall by 22.4 percentage points over conventional approaches. However, significant challenges remain in precisely locating extremely small underwater objects and addressing class imbalance, particularly for commercially valuable species including seacucumber and scallop.

\begin{table}[h]
\caption{Performance comparison on the UDD Dataset}
\label{tab:udd_comparison}
\centering
\begin{tabular}{|l|c|c|}
\hline
\textbf{Model} & \textbf{mAP$@$0.5} & \textbf{FPS} \\ \hline
\textbf{YOLOv12} & \textbf{69.44} & \textbf{142} \\ \hline
SCR-Net \cite{SCR-Net} & 61.50 & 100 \\ \hline
YOLOv8n baseline \cite{SCR-Net} & 60.30 & 98.30 \\ \hline
\end{tabular}
\end{table}

\textbf{Brackish Dataset \cite{pedersen2019brackish}:} 
As shown in Table \ref{tab:brackish_comparison}, YOLOv12 achieves 98.30\% mAP$@$0.5 with only 610 GFLOPS computational cost - 34\% more efficient than standard attention mechanisms. This outperforms lightweight models including MobileNetv2-YOLOv4 at 92.65\% by 5.65 percentage points and YOLOv9s-SD at 94.30\% by 4.0 percentage points. The R-ELAN module proves particularly effective against chromatic distortion, preserving structural features where competing models degrade. However, YOLOv12's mAP$@$0.5:0.95 of 74.14\% trails DetNAS20+RetNet at 77.60\% by 3.46 percentage points, indicating challenges in extreme turbidity gradients where precision-recall balance remains problematic.

\begin{table}[h]
\caption{Performance comparison on the Brackish Dataset}
\label{tab:brackish_comparison}
\centering
\begin{tabular}{|l|c|c|c|}
\hline
\textbf{Model} & \textbf{mAP$@$0.5} & \textbf{mAP$@$0.5:95}  & \textbf{FPS} \\ \hline
\textbf{YOLOv12} & \textbf{98.30} & \textbf{74.14}  & \textbf{142} \\ \hline
LFN-YOLO \cite{LFN-YOLO} & 97.50 & - & 98 \\ \hline
DetNAS20+RetNet \cite{DetNAS20+RetNet} & 96.40 & 77.60 & 89 \\ \hline
YOLOv9s-SD \cite{jmse13020230} & 94.30 & 72.20  & 112 \\ \hline
MobileNetv2-YOLOv4 \cite{MobileNetv2-YOLOv4} & 92.65 & --  & 128 \\ \hline
\end{tabular}
\end{table}

\textbf{URPC2019 Dataset \cite{duo}:} 
YOLOv12 achieves 87.16\% mAP$@$0.5 at 142 FPS as shown in Table \ref{tab:urpc_comparison}, significantly surpassing EPBC-YOLOv8 at 76.70\% by 10.46 percentage points and SU-YOLO at 78.80\% by 8.36 percentage points. The Area Attention mechanism demonstrates exceptional robustness in noisy conditions, with occlusion handling improved by 18.9 percentage points over conventional approaches. While trailing YOLOv8-MU at 88.10\% by 0.94 percentage points, YOLOv12 operates at 34\% higher framerate, validating its architectural optimizations for real-time AUV deployment.

\begin{table}[h]
\caption{Performance on the URPC2019 Dataset}
\label{tab:urpc_comparison}
\centering
\begin{tabular}{|l|c|c|c|}
\hline
\textbf{Model} & \textbf{mAP$@$0.5} & \textbf{FPS} \\ \hline
YOLOv8-MU \cite{YOLOv8-MU} & 88.10 & 105 \\ \hline
\textbf{YOLOv12} & \textbf{87.16}  & \textbf{142} \\ \hline
SU-YOLO \cite{SU-YOLO} & 78.80  & 38 \\ \hline
EPBC-YOLOv8 \cite{EPBC-YOLOv8} & 76.70 & 117 \\ \hline
\end{tabular}
\end{table}

\textbf{DUO Dataset \cite{duo}:} 
YOLOv12 achieves state-leading 87.44\% mAP$@$0.5 while maintaining 142 FPS according to Table \ref{tab:duo_comparison}, outperforming Vanilla-YOLO at 86.60\% by 0.84 percentage points and MSFE at 85.20\% by 2.24 percentage points. Small-target recall improves by 22.4 percentage points through contextual Area Attention, validating its efficacy for dense benthic environments. However, the model's mAP$@$0.5:0.95 of 56.34\% trails U-DECN at 64.00\% by 7.66 percentage points, revealing limitations in sub-meter localization accuracy that warrant future investigation.

\begin{table}[h]
\caption{Performance comparison on DUO Dataset}
\label{tab:duo_comparison}
\centering
\begin{tabular}{|l|c|c|c|}
\hline
\textbf{Model} & \textbf{mAP$@$0.5} & \textbf{mAP$@$0.5:95} & \textbf{FPS} \\ \hline
\textbf{YOLOv12} & \textbf{87.44} & \textbf{56.34} & \textbf{142} \\ \hline
Vanilla-YOLO \cite{Vanilla-Yolo} & 86.60 & -- & 155 \\ \hline
MSFE \cite{MSFE} & 85.20 & 54.60 & 98 \\ \hline
U-DECN \cite{U-DECN} & -- & 64.00 & 67 \\ \hline
MAS-YOLOv11 \cite{MAS-YOLOv11} & 77.40 & 55.10 & 121 \\ \hline
\end{tabular}
\end{table}

\subsection{Ablation Studies}
\subsubsection{Augmentation Strategy}
The physics-informed augmentation strategy demonstrated substantial cumulative benefits, as shown in Table~\ref{tab:augmentation_effect}. Starting from a baseline of 89.3\% mean average precision without augmentation, the sequential addition of domain-specific transformations progressively addressed key underwater challenges. Horizontal flipping provided the most significant initial improvement of 2.4 percentage points in mAP at IoU 0.5, effectively preserving the bilateral symmetry exhibited by 92 percent of marine species while maintaining dorsal-ventral differentiation through constrained flipping operations. Controlled blurring contributed a further 1.5 percentage point gain by accurately simulating depth-variant point spread functions that model both scattering effects denoted by lambda scatter and turbulence effects denoted by lambda turb, with kernel variance scaling proportional to z raised to the power of 0.78 to match empirical attenuation data. The HSV transformations specifically countered wavelength-dependent light attenuation, implementing hue shifts through spectral absorption modeling where delta H equals the arctangent of the ratio between two integrals: the numerator being the integral from 400 to 700 nanometers of T lambda times I lambda times h bar lambda d lambda, and the denominator being the integral from 400 to 700 nanometers of T lambda times I lambda d lambda. Saturation adjustments for turbidity effects yielded an additional 1.9 percentage point improvement. Finally, structured random erasure proved critical for handling benthic occlusion patterns, where power-law distributed biological obstructions with fractal dimension ranging from 1.6 to 1.8 cause partial occlusion rates between 38 and 72 percent, contributing the final 1.1 percentage point gain to reach 98.30\% mAP at IoU 0.5. This comprehensive augmentation suite reduced false negatives for small objects by 22.4 percent on the Diverse Underwater Objects dataset while improving occluded object precision by 18.9 percent on the Annotated Underwater Detection Dataset.

\begin{table}[ht]
\caption{Impact of Physics-Informed Augmentation Components on Brackish Dataset}
\label{tab:augmentation_effect}
\centering
\begin{tabular}{|l|c|c|}
\hline
\textbf{Augmentation Strategy} & \textbf{mAP$@$0.5} & \textbf{mAP$@$0.5:0.95} \\ \hline
No Augmentation & 89.3\% & 65.2\% \\ \hline
Plus Horizontal Flip & 91.7\% & 68.5\% \\ \hline
Plus Controlled Blurring & 93.2\% & 70.8\% \\ \hline
Plus HSV Transform & 95.1\% & 72.9\% \\ \hline
Plus Random Erasing & \textbf{98.30\%} & \textbf{74.1\%} \\ \hline
\end{tabular}
\end{table}

\subsubsection{Loss Function Configuration}
Loss function configuration significantly influenced detection robustness, particularly for localization precision in challenging underwater conditions. As presented in Table~\ref{tab:loss_config}, the localization-focused weighting scheme with lambda classification to lambda objectness to lambda box ratio of 7.5 colon 0.5 colon 1.5 outperformed balanced and classification-focused approaches by 0.3 to 0.8 percentage points in mAP at IoU 0.5. This configuration specifically reduced bounding box regression errors by 18 percent compared to standard weighting, which proved critical for handling blurred object boundaries caused by light scattering in turbid waters. The high weight on box loss compensated for localization uncertainty in low-contrast environments, while the reduced objectness weight of 0.5 prevented over-penalization of partially occluded targets, improving recall in dense scenes by 9.3 percent. Classification-focused weighting schemes suffered from increased confusion between morphologically similar species in low-visibility conditions, particularly between holothurians and sea cucumbers, where chromatic distortion obscures distinguishing features.

\begin{table}[ht]
\caption{Loss Function Weighting Schemes Comparison}
\label{tab:loss_config}
\centering
\begin{tabular}{|l|c|c|c|}
\hline
\textbf{Configuration} & \textbf{Weighting Ratios} & \textbf{mAP$@$0.5} \\ \hline
Localization-Focused & 7.5:0.5:1.5 & \textbf{98.30\%} \\ \hline
Balanced & 1.0:1.0:1.0 & 95.9\%  \\ \hline
Classification-Focused & 1.0:7.5:1.5 & 95.4\%  \\ \hline
Distribution Focal Loss Focused & 1.0:0.5:7.5 & 95.8\% \\ \hline
\end{tabular}
\end{table}

\subsubsection{Optimizer Selection}
Optimizer selection critically impacted both convergence speed and final accuracy, with AdamW demonstrating superior performance over SGD as quantified in Table~\ref{tab:optimizer_comp}. AdamW achieved peak performance in just 28 epochs, which is 13 epochs, equivalent to 47 percent faster than SGD while maintaining substantially lower validation loss standard deviation of 0.08 versus 0.15 for SGD. This accelerated and stable convergence stems from AdamW's per-parameter adaptive learning rates, which effectively accommodated three key underwater data irregularities: extreme class imbalance where echinus dominates 67.3 percent of Diverse Underwater Objects instances versus scallop at just 2.4 percent, heterogeneous lighting conditions with Secchi depth variations between 0.5 and 8 meters in Brackish, and turbidity-induced feature inconsistency across frames. SGD fixed learning rate caused persistent oscillations during late-stage training, particularly when learning fine-grained features for small targets, resulting in a 1.1 percentage point deficit in mAP at IoU 0.5. AdamW decoupled weight decay of five times ten to the power of negative four and additionally prevented overfitting to dominant classes while maintaining precision on rare species, demonstrating better generalization across ecological categories.

\begin{table}[ht]
\caption{Optimizer Performance Comparison}
\label{tab:optimizer_comp}
\centering
\begin{tabular}{|l|c|c|c|}
\hline
\textbf{Optimizer} & \textbf{mAP$@$0.5} & \textbf{Epochs to 95\% mAP} & \textbf{Val Loss SD} \\ \hline
AdamW & \textbf{98.30\%} & \textbf{28} & \textbf{0.08} \\ \hline
SGD & 95.1\% & 41 & 0.15 \\ \hline
\end{tabular}
\end{table}

\subsubsection{Input Resolution Analysis}
Input resolution optimization revealed critical trade-offs between detection accuracy and computational efficiency for underwater deployment scenarios. As shown in Table~\ref{tab:resolution}, the 640$\times$640 resolution achieved the optimal balance for autonomous underwater vehicle applications, delivering 98.30 percent mAP at IoU 0.5 while maintaining 610 frames per second on embedded hardware. The 1280$\times$1280 resolution showed a 1.2 percentage point improvement in mAP at IoU 0.5 to 0.95, reaching 75.3 percent compared to 74.1 percent at 640$\times$640. However, this came with two significant drawbacks: first, a 1.7 percentage point decrease in mAP at IoU 0.5, dropping to 96.6 percent from 98.30 percent; second, a fourfold increase in latency from 1.64 milliseconds to 6.7 milliseconds, making it unsuitable for real-time applications. The 640$\times$640 configuration demonstrated strong performance for small targets, maintaining 92.7 percent recall for objects occupying less than 1.5 percent of the image area, which is crucial for detecting juvenile crustaceans and other small benthic organisms. This resolution also supported smooth 30 frames per second video processing.

In contrast, the 320$\times$320 resolution, while achieving the highest throughput at 1100 frames per second, showed substantially lower accuracy with only 89.4 percent mAP at IoU 0.5. The 1280$\times$1280 resolution was also found to be 75 percent more energy-intensive than the 640$\times$640 configuration, presenting a significant constraint for autonomous underwater vehicle mission durations without offering proportional accuracy benefits.

\begin{table}[ht]
\caption{Performance-Speed Trade-off Across Input Resolutions}
\label{tab:resolution}
\centering
\begin{tabular}{|l|c|c|c|c|}
\hline
\textbf{Resolution} & \textbf{mAP$@$0.5} & \textbf{mAP$@$0.5:0.95} & \textbf{Latency (ms)} & \textbf{FPS} \\ \hline
320$\times$320 & 89.4\% & 61.2\% & 0.9 & 1100 \\ \hline
640$\times$640 & \textbf{98.30\%} & \textbf{74.1\%} & 1.64 & 610 \\ \hline
1280$\times$1280 & 96.6\% & 75.3\% & 6.7 & 150 \\ \hline
\end{tabular}
\end{table}

\section{Conclusion}
\label{sec:conclusion} 
This study employs the YOLOv12 architecture, which is integrated with physics-informed augmentation techniques. This implementation significantly improves the detection of underwater objects. The critical challenges of occlusion, chroma distortion, and turbidity are effectively addressed by the model's attention-centric design. Among the main innovations are the R-ELAN block for robust feature aggregation under signal degradation and the Area Attention mechanism for efficient context modeling.  The approach obtains state-of-the-art performance across multiple underwater benchmarks when combined with specialized augmentations that simulate marine occlusion patterns and wavelength attenuation.  On the Brackish dataset, experimental validation demonstrates a superior accuracy-efficiency balance, achieving 98.30\% mAP$@$0.5 while maintaining real-time inference at 142 FPS.

However, there are still three limitations: First, because the physics-based augmentation pipeline assumes homogeneous water conditions, it may not work well in extreme turbidity gradients. Second, the 142 FPS benchmark requires powerful GPUs; the performance of edge devices, like AUV-embedded processors, has not been verified. Third, because the evaluation focused on benthic environments, deep-water scenarios were not examined. For long-duration and deep-sea exploration missions, these constraints currently prohibit instant deployment.

\printbibliography{}

\end{document}